\documentclass[conference,a4paper,10pt]{IEEEtran}
\IEEEoverridecommandlockouts

\usepackage{cite}
\usepackage{amsmath,amssymb,amsfonts}
\usepackage{algorithm}
\usepackage{algorithmic}
\usepackage[dvipdfmx]{graphicx}
\usepackage{bmpsize}
\usepackage{textcomp}
\usepackage{xcolor}
\def\BibTeX{{\rm B\kern-.05em{\sc i\kern-.025em b}\kern-.08em
    T\kern-.1667em\lower.7ex\hbox{E}\kern-.125emX}}
\usepackage{times}
\usepackage{epsfig}
\usepackage{subfigure}

\makeatletter
\IEEEtriggercmd{\reset@font\normalfont\fontsize{10pt}{10pt}\selectfont}
\makeatother
\IEEEtriggeratref{1}

\restylefloat{algorithm}

\DeclareMathAlphabet{\mathpzc}{OT1}{pzc}{m}{it}

\newcommand{\argmin}[2]{\underset{#1}{\mathrm{argmin}}\bigg\{#2\bigg\}}

\usepackage[breaklinks=true,bookmarks=false]{hyperref}

\begin{document}
	
	\title{Incorporating the Barzilai-Borwein Adaptive Step Size into Sugradient Methods for Deep Network Training}
	
	\author{Antonio Robles-Kelly \ \ \ \ \ \  \ \ \ \ \ \ Asef Nazari\\
		School of IT, Faculty of Eng., Sci. and Build Env., Deakin University, Waurn Ponds, VIC 3216, Australia\\
	}

	\maketitle
	\begin{abstract}
		In this paper, we incorporate the Barzilai-Borwein\cite{barzilai:88} step size into gradient descent methods used to train deep networks. This allows us to adapt the learning rate using a two-point approximation to the secant equation which quasi-Newton methods are based upon. Moreover, the adaptive learning rate method presented here is quite general in nature and can be applied to widely used gradient descent approaches such as Adagrad\cite{duchi:2011} and RMSprop. We evaluate our method using standard example network architectures on widely available datasets and compare against alternatives elsewhere in the literature. In our experiments, our adaptive learning rate shows a smoother and faster convergence than that exhibited by the alternatives, with better or comparable performance.
	\end{abstract}
	



\maketitle


\section{Introduction}
The learning rate is a hyperparameter that, in neural network architectures, controls the influence of the loss gradient upon the update of the weights in the net. This is a common approach in the update step of gradient descent methods, where its often referred in the optimisation literature as the ``step size''\cite{kim:2016}. Indeed, the step size has been thoroughly studied in the optimisation community, leading to important results such as those by Barzilai and Borwein \cite{barzilai:88} and Nesterov \cite{nesterov:2003}.

In the machine learning literature, the learning rate has attracted attention due to its relevance to the training of neural networks. Along these lines, Senior {\it et al.} \cite{senior:2013} have studied the learning rate and its effect in speech recognition. In \cite{senior:2013}, empirical evidence is provided so as to sustain the notion that a careful choice of learning rate does improve the training results and convergence. This is not surprising, however, since this is also well known in the optimisation community regarding the step size of gradient descent methods.

Indeed, the learning rate is known to dramatically affect the model obtained after training \cite{goodfellow:2016}. Unfortunately, there is no analytic means to set the learning rate. Moreover, since the learning rate becomes subsumed in the chain rule on the update process for each layer of the network as the back propagation scheme advances, it is known to be highly non-linear in behaviour. 

As a result, there are an ample set of schemes and techniques to adjust the training rate for deep neural networks. Maybe the most commonly used one is the use of a ``schedule'' so as to adjust the learning rate on a per-epoch scheme, {\it i.e.} over time. This approach includes constant, running average and search-then-converge schemes \cite{darken:91}. Despite effective, these schemes are often based on a careful set of the decay parameters and greatly depend upon the number of epochs, size of the training dataset and depth of the network. These also often apply the learning rate per-epoch, {\it i.e.} over time, for all layers in the network, regardless of their position or value of the loss gradient.

Further, adaptive learning rate schemes have been explored since early on in the neural networks literature. For instance, in \cite{jacobs:88}, Jacobs investigates the use of an adaptive learning rate for improving rates of convergence in steepest descent back propagation, where he asserts that the learning rate should be allowed to vary over time and per-weight in the network.  Adagrad \cite{duchi:2011}, for instance, employs a learning rate given by the square root of sum of squares of the gradient for past epochs. Due to its simplicity, Adagrad also has the advantage of having a low computational burden and being easy to implement. 

Despite these advantages, Adagrad has the drawback of tending to yield diminishing updates as it progresses. This is due to the sum of squares of the gradients involved in its computation, which may derive in situations where the training does not fully converge. As a result, Root Mean Square propagation (RMSprop) \footnote{RMSprop, to our knowledge is unpublished. Nonetheless, its widely used in the machine learning community.Its description can be found at \url{http://www.cs.toronto.edu/~tijmen/csc321/slides/lecture_slides_lec6.pdf}} and Adadelta \cite{zeiler:2018} restrict the number of past gradients used for the computation of the sum of squares either by using a scheme akin to momentum \cite{qian:99} or by applying an exponentially decaying average. Adaptive moment estimation (Adam) \cite{kingma:2015}, in the other hand, employs the mean and variance of past gradients so as to derive an update rule similar to that of Adadelta and RMSprop. 

Similarly, Nesterov-accelerated adaptive moment estimation (Nadam) \cite{dozat:2016}, modifies the update of the methods above to compute the gradient after the momentum-based step has been added, not before. AMSGrad \cite{reddi:2018} tackles the problem of an increasing step-size sometimes exhibited by Adam making use of the maximum for the past squared gradients over the update schedule. 

In this paper, we explore the inclusion of the Barzilai-Borwein\cite{barzilai:88} adaptive step size into the gradient descent optimisation used in the training of deep networks. To this end, we depart from the relationship between gradient descent and the learning rate so as to incorportate the Barzilai-Borwein\cite{barzilai:88} step size formulation in subgradient methods. We show how this learning rate can be used, in a straightforward manner, in conjuction with methods such as Adagrad\cite{duchi:2011} and RMSProp to train a deep network. We perform experiments using standard example network architectures using MatConvNet\cite{vedaldi:2015} on CIFAR10, Imagenet and MNIST. In our experiments, our adaptive step size shows a smoother and faster convergence than that exhibited by Adam\cite{dozat:2016} and Adadelta\cite{zeiler:2018}.

\section{Learning Rate Adaptation}

\subsection{Background}

As mentioned above, training of deep networks is often effected using gradient descent. In a neural network, the aim of computation at training is the minimisation of a loss, or cost, function $\mathcal{L}(\Theta,y^\prime,y)$. That is, we aim at recovering the parameters such that
\begin{equation}
\begin{split}
\Theta^*&=\argmin{\Theta}{\mathcal{L}(\Theta,y^\prime,y)}\\\nonumber
&=\argmin{\Theta}{L(y^\prime,y)+\lambda\mathcal{R}(\Theta)}
\end{split}
\end{equation}
where the objective function $L(y^\prime,y)$ depends on both, the prediction $y^\prime$ and the target $y$ for each of the input instances $x^0$ in the training set, $\mathcal{R}(\Theta)$ is a regularisation term that depends on the parameters $\Theta$ and $\lambda\geq 0$ is a weight that controls the influence of the regulariser in the minimisation above.

It is worth noting that, nonetheless in deep networks the parameters are given by the weights in the connections. In the equations above and throughout the paper we make no distinction between these in fully connected and convolutional layers. We have done this for the sake of generality since the developments in this and the following section are quite general in nature and apply to these indistinctively.   


To obtain the gradient for the loss function, we can employ the chain rule and write
\begin{equation}
   \nabla_\theta L(\theta,y^\prime,y)= \nabla_{y^\prime} L(y^\prime,y)\nabla_{\theta}y^\prime
\end{equation}

Moreover, let the evaluation function at the layer indexed $l$ in the network with $N$ layers be $J^{(l)}(\theta^{l},x^{l})$ with parameters $\theta^{l}\in\Theta$. By taking the prediction to be the value of the evaluation at the last, {\it i.e.} $N^{th}$, layer, we can write 
\begin{equation}
   \nabla_{\theta^N} \mathcal{L}(\theta,y^\prime,y)= \nabla_{y^\prime} L(y^\prime,y)\nabla_{\theta^{N}}J^{(N)}(\theta^{N},x^{N})+\nabla_{\theta^{N}}\mathcal{R}(\Theta)
\end{equation}
where we have used the shorthand $y^\prime=J^{(N)}(\theta^{N},x^{N})$.

This is indeed the basis of the back propagation algorithm for training 
nerual networks. Note the prediction $y^\prime$ can be written as follows
\begin{equation}
    y^\prime=J^{(N)}(\theta^{N},x^{N})\circ J^{(N-1)}(\theta^{N-1},x^{N-1})
\end{equation}
where $\circ$ is the composition operator.

This is important since, using the chain rule, we can now write
{\small
\begin{equation}
   \nabla_{\theta^{N-1}} y^\prime=
   \nabla_{\theta^{N}}J^{(N)}(\theta^{N},x^{N}) \nabla_{\theta^{N-1}}J^{(N-1)}(\theta^{N-1},x^{N-1})
\end{equation}}
and, thus, for the layer indexed $l$ in the network, we have
\begin{equation}
    \nabla_{\theta^{l}} y^\prime=\prod_{i=0}^{N-l}\nabla_{\theta^{N-i}}J^{(N-i)}(\theta^{N-i},x^{N-i})
    \label{eq:05}
\end{equation}

This is the basis for the backpropagation step during the training process in neural networks, where the gradient is used to update the parameter set of the network layer-by-layer, starting from the last to the first, {\it i.e.} from the output layer to the input one. 

\subsection{Gradient Descent and Learning Rates}

As mentioned above, the training of the network is often effected making using gradient descent. This is done by updating the parameters $\theta^l$ of the functions $J^l(\theta^l,x^l)$ making use of the rule
\begin{equation}
  \theta^l_{k+1}=\theta^l_{k}-\alpha_k G_k(\theta^l_k)
  \label{eq:06}
\end{equation}
where $\theta^l_{k}$ is the $k^{th}$ iterate of $\theta^l$, $G_k(\theta^l_k)$ is one of the sub-gradients of $\mathcal{L}(\theta,y^\prime,y)$ at $\theta^l_{k}$ and $\alpha_k$ is the step size, {\it i.e.} learning rate. 

Since the loss function is differentiable, we can set
\begin{equation}
    G_k(\theta^l_k)=\big[\nabla_{y^\prime} L(y^\prime,y)\nabla_{\theta^{l}} y^\prime+\nabla_{\theta^{l}}\mathcal{R}(\Theta)\big]_{k}
\end{equation}
where $\nabla_{\theta^{l}} y^\prime$ is given by Equation \ref{eq:05} and we have written $[\cdot]_k$ to imply the evaluation of the gradient at the iterate $k$.

As mentioned previously, the learning rate $\alpha_k$ can be set using a number of strategies. The simplest of these is that of using a constant, this is $\alpha_k=\eta$ or a decreasing scheme such as 
\begin{equation}
    \alpha_k=\frac{b}{\sqrt{k}}
    \label{eq:07}
\end{equation}
where $b$ is a bound, real constant. 

Another popular option is to use an exponential decay. In the case of Adagrad \cite{duchi:2011}, this is done per-parameter, whereby Equation \ref{eq:06} takes the form 
\begin{equation}
    \theta^l_{k+1}=\theta^l_{k}-\mathbf{a}_k\odot G_k(\theta^l_k)
    \label{eq:08}
\end{equation}
where $\odot$ denotes the elementwise product and $\mathbf{a}_k$ is a vector whose entry indexed $j$ is given by
\begin{equation}
    a_{j}=\frac{b}{\sqrt{\mathcal{C}_{j}+\epsilon}}
\end{equation}
where $\epsilon$ is a small, real-valued constant that avoids division by zero errors and $C_{j}$ is the sum of squares of the subgradient $G_k(\theta^l_k)$ for the $j^{th}$ parameter up to the $k^{th}$ iteration.

It is worth noting, that, in practice, a momentum \cite{qian:99} term is added to the update so as to accelerate the convergence of the optimisation process. The addition of momentum is effected by using the values of the subgradient over the past iterations of the method as an alternative to the direct application of $G_k(\theta^l)$ in Equation \ref{eq:08}. Thus, we employ the following update
\begin{equation}
  \theta^l_{k+1}=\theta^l_{k}-\mathbf{a}_k\odot \mathcal{V}_k(\theta^l_k)
  \label{eq:10}
\end{equation}
where
\begin{equation}
    \mathcal{V}_{k}(\theta^l_k)=\beta\mathcal{V}_{k-1}(\theta^l_{k-1})+(1-\beta)G_k(\theta^l_k)
    \label{eq:11}
\end{equation}
and $\beta\in[0,1]$

\subsection{Incorporating the Barzilai-Borwein Step Length}

We now turn our attention to the Barzilaai-Borwein approach gradient descent method. This approach, published in \cite{barzilai:88} is a Hessian-free optimisation method motivated by Newton's method \cite{bonnans:2006}. The Barzilai-Borwein method employs a particular step size which, with low extra computational cost, often delivers a noticeable increase in performance with respect to traditional gradient descent approaches employing the update rule in Equation \ref{eq:06}.

The idea underpinning the Barzilai-Borwein method stems from the notion that Newton's method does incorporate second order information with the disadvantage of the need for the computation of the Hessian. To avoid the direct computation of the Hessian matrix, Barzilai and Borwein propose to use a step size that approximates its inverse magnitude.

Making use of the notation in the previous section, consider, as an alternative to that in Equation \ref{eq:08}, the following parameter update rule
\begin{equation}
  \theta^l_{k+1}=\theta^l_{k}-\mathbf{a}_k\odot\big[\mathbf{H}\mathcal{L}^l(\theta,y^\prime,y)\big]^{-1} G_k(\theta^l_k)
  \label{eq:12}
\end{equation}
where $\mathbf{H}\mathcal{L}^l(\theta,y^\prime,y)$ denotes the Hessian  of the loss function with respect to the parameters $\theta_k^l$ of the layer indexed $l$ at the corresponding iterate.

Let $\Delta \theta^l_k=\theta^l_k-\theta^l_{k-1}$. The Barzilai and Borwein approach \cite{barzilai:88} assumes the the Hessian $\mathbf{H}\mathcal{L}^l(\theta,y^\prime,y)$ satisfies the secant equation, {\it i.e.}
\begin{equation}
  \Delta\theta^l_{k}=\mathbf{a}_k\odot\big[\mathbf{H}\mathcal{L}^l(\theta,y^\prime,y)\big]^{-1} \Delta G_{k}(\theta^l_{k})
  \label{eq:14}
\end{equation}
where $\Delta G_{k}(\theta^l_{k})=G_{k}(\theta^l_{k})-G_{k-1}(\theta^l_{k-1})$.

Barzilai and Borwein \cite{barzilai:88} then use a least-squares solution of Equation \ref{eq:14} such that
the step size (learning rate) at the layer indexed $l$ at iteration $k$ corresponds to the following minimisation
\begin{equation}
    \gamma^l_k = \argmin{\beta}{
    \Arrowvert\Delta\theta^l_{k}-\mathbf{a}_k\odot\beta G_{k}(\theta^l_{k})\Arrowvert}
\end{equation}

By treating $\Delta\theta^l_{k}$ and $\Delta G_{k}(\theta^l_{k})$ as vectors, we can write the solution as follows
\begin{equation}
    \gamma^l_k = \frac{\big[\Delta\theta^l_{k}\big]^T\Delta G_{k}(\theta^l_{k})}{\big[\Delta G_{k}(\theta^l_{k})\big]^T \Delta G_{k}(\theta^l_{k})}
    \label{eq:16}
\end{equation}
where, as usual, $\big[\cdot\big]^T$ denotes the transpose operation.

Making use of the approximation $\gamma^l_k$ to the Hessian, we can use the update
\begin{equation}
  \theta^l_{k+1}=\theta^l_{k}-\gamma^l_k\big(\mathbf{a}_k\odot G_k(\theta^l_k)\big)
  \label{eq:17}
\end{equation}
as an alternative to that in Equation \ref{eq:12}.

\section{Implementation and Results}

For our implementation we use MatConvNet \cite{vedaldi:2015}\footnote{MatConvNet is widely accessible at \url{http://www.vlfeat.org/matconvnet/}}. The main reasons for this choice of implementation platform are twofold. Firtly, MatConvNet provides a means for a set of widely available network architectures that can be trained using stochastic gradient descent methods on standard datasets, {\it i.e.}  MNIST, CIFAR10,  and ImageNet. Secondly, MatConvNet is distributed with standard solver options for Adadelta\cite{zeiler:2018}, Adagrad\cite{duchi:2011}, Adam\cite{dozat:2016} and RMSProp.

In Algorithm \ref{alg:01} we show the pseudocode for our Barzilai-Borwein learning rate computation. For the sake of computational ease, we have used notation akin to that used in the equations in the previous sections. That said, there are a few aspects to note in the algorithm. Firstly, throughout the algorithm, we have dropped the indices for the iteration and layer number. We have done this for the sake of clarity since the algorithm and the corresponding variables apply per-layer at each iterate.

Algorithm \ref{alg:01} requires at input the gradients $G_{new}$ and $G_{old}$ for the current and previous iterations and updates the variable $\Delta\theta$ at each call, updating the parameters before returning them in Line 5. The algorithm also requires a constant $\epsilon$, which we have introduced so as to avoid divisions by zero induced by the term $[G_{k}(\theta^l_{k})\big]^T \Delta G_{k}(\theta^l_{k})$ in Equation \ref{eq:16} when its very close to convergence. For our method, we have set $\epsilon=1\times 10^{-8}$. 

Also, note that in Algorithm \ref{alg:01} we employ $C$ to denote the squared sum of gradients used by Adagrad\cite{duchi:2011}. This applies equally to both, Adagrad\cite{duchi:2011} and RMSprop. To employ RMSprop instead of Adagrad\cite{duchi:2011} in Algorighm \ref{alg:01}, Line 3 should be substituted accordingly. That said, Adagrad\cite{duchi:2011} and RMSprop employ the same function arguments in MatConvNet, which makes the substitution a straightforward one.

At the very first iteration of the learning process the variable $G_{old}$ in Algorithm \ref{alg:01} is not available. As a result, we initialise the training calling Adam\cite{dozat:2016}. After the first iteration, we can apply Algorithm \ref{alg:01} in a straightforward manner. In all our experiments, we have used the default value of epsilon and rho, which are $\epsilon=1\times 10^{-10}$ and $\rho=1$ for Adagrad\cite{duchi:2011} and $\epsilon=1\times 10^{-8}$ and $\rho=0.99$ for RMSprop. 

For our experiments, we have used the example convolutional networks provided with MatConvNet v1.0-beta25 for Imagenet, MNIST and CIFAR10. All our results were computed using Matlab R2018a running on a workstation with a Testal K40c GPU.

\begin{algorithm}[t]
 \caption{Barzilai-Borwein learning rate computation}
 \begin{algorithmic}[1]
 \renewcommand{\algorithmicrequire}{\textbf{Input:}}
 \renewcommand{\algorithmicensure}{\textbf{Output:}}
 \REQUIRE $\theta_{old}$, $\Delta\theta$, $G_{old}$, $G_{new}$, $\mathbf{C},\epsilon$
 \ENSURE  $\theta_{new}$, $\Delta\theta$, $\mathbf{C}$
  \STATE $\Delta G = G_{new}-G_{old}$
  \STATE $\gamma = \frac{\big[\Delta\theta\big]^T\Delta G}{\big[\Delta G\big]^T \Delta G+\epsilon}$
  \STATE $[\theta_{new},\mathbf{C}]$ = Adagrad$(\theta_{old}, \mathbf{C}, G_{new}, \gamma)$
  \STATE $\Delta\theta = \theta_{new}-\theta_{old}$
 \RETURN  $\theta_{new}$, $\Delta\theta$, $\mathbf{C}$
 \end{algorithmic} 
 \label{alg:01}
 \end{algorithm}

\begin{figure*}
	\centering
	\includegraphics[width = 0.32\textwidth,height = 0.23 \textheight]{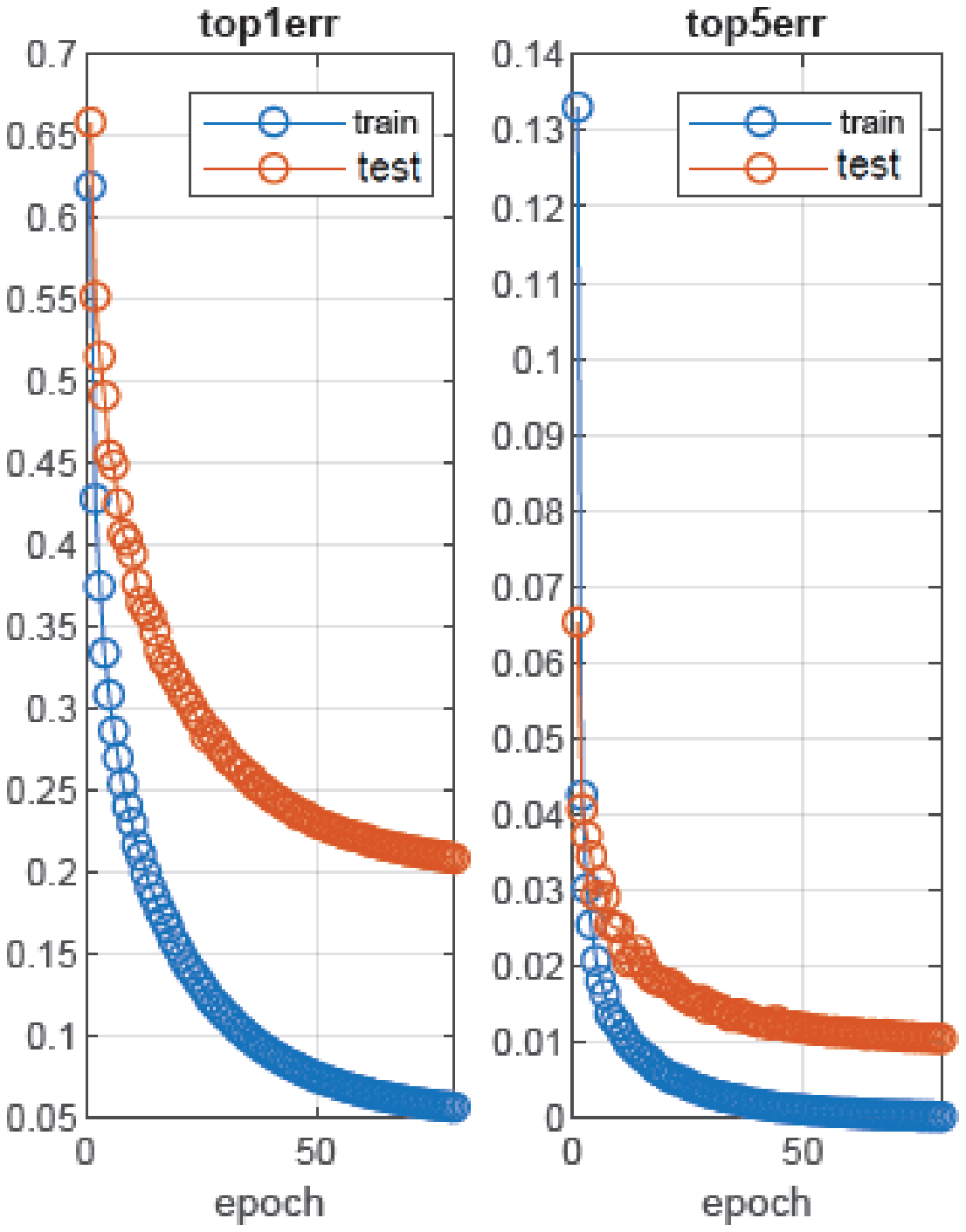}
	\includegraphics[width = 0.32\textwidth,height = 0.23 \textheight]{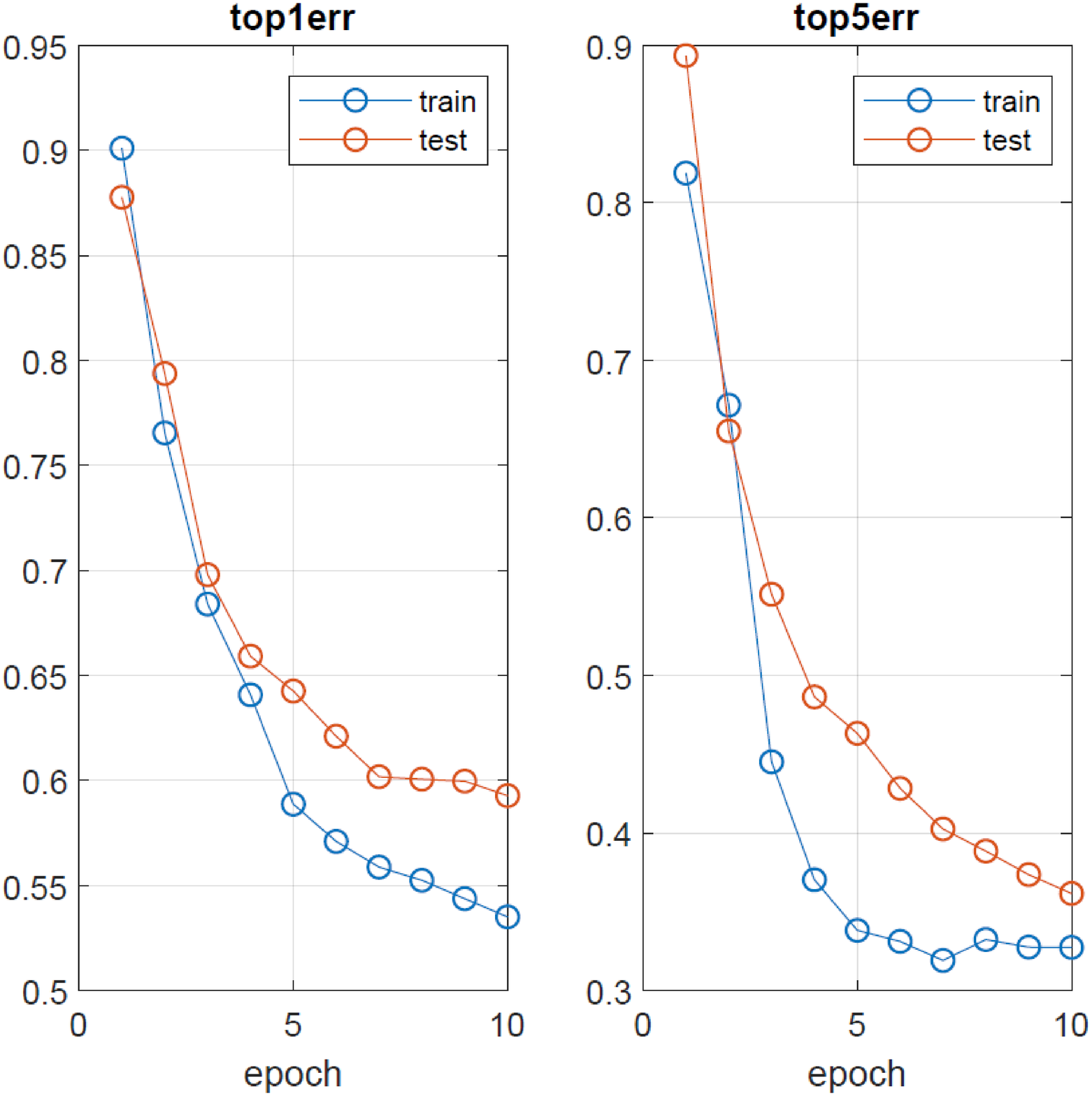}
	\includegraphics[width = 0.32\textwidth,height = 0.23 \textheight]{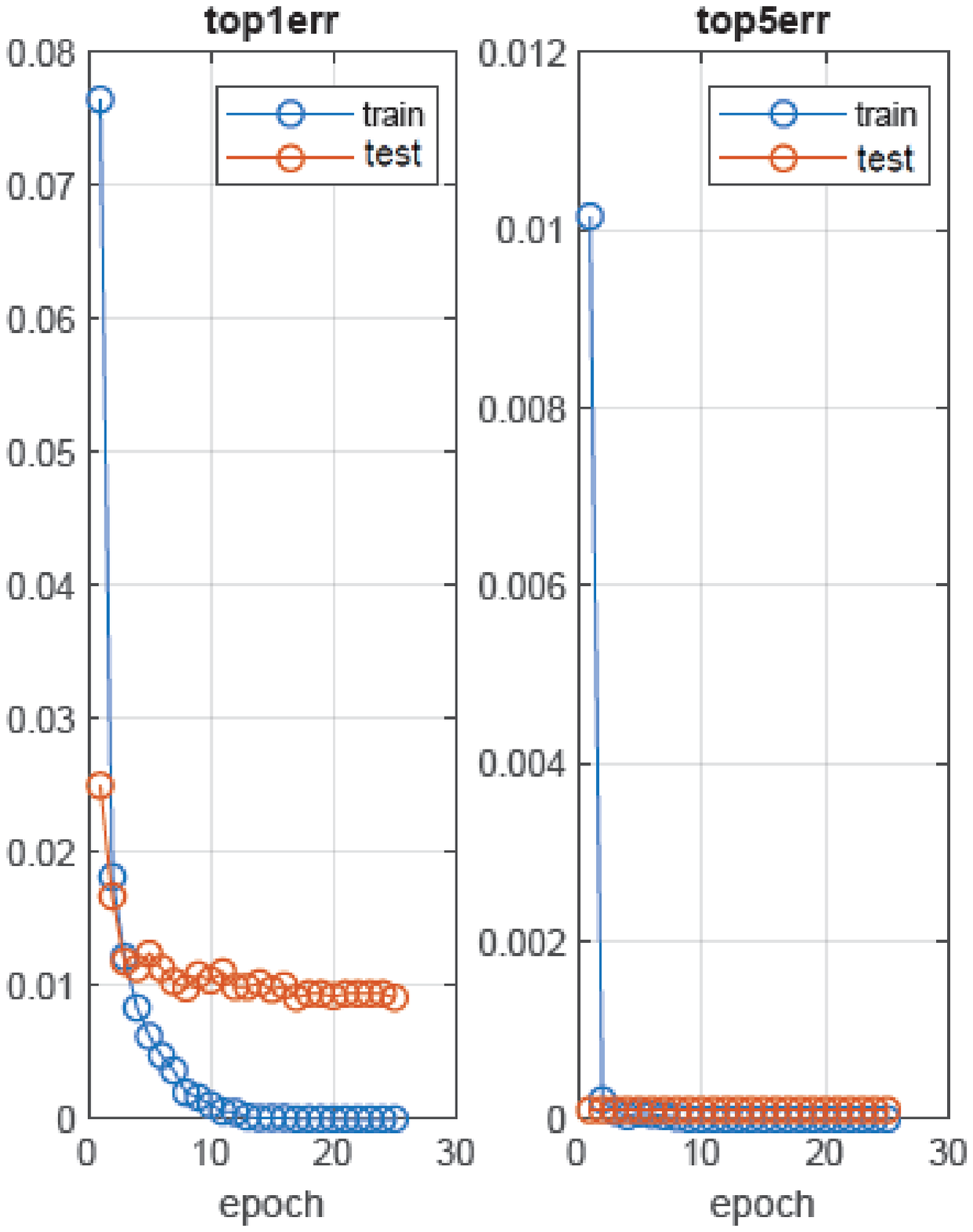}\\
	\includegraphics[width = 0.32\textwidth,height = 0.23 \textheight]{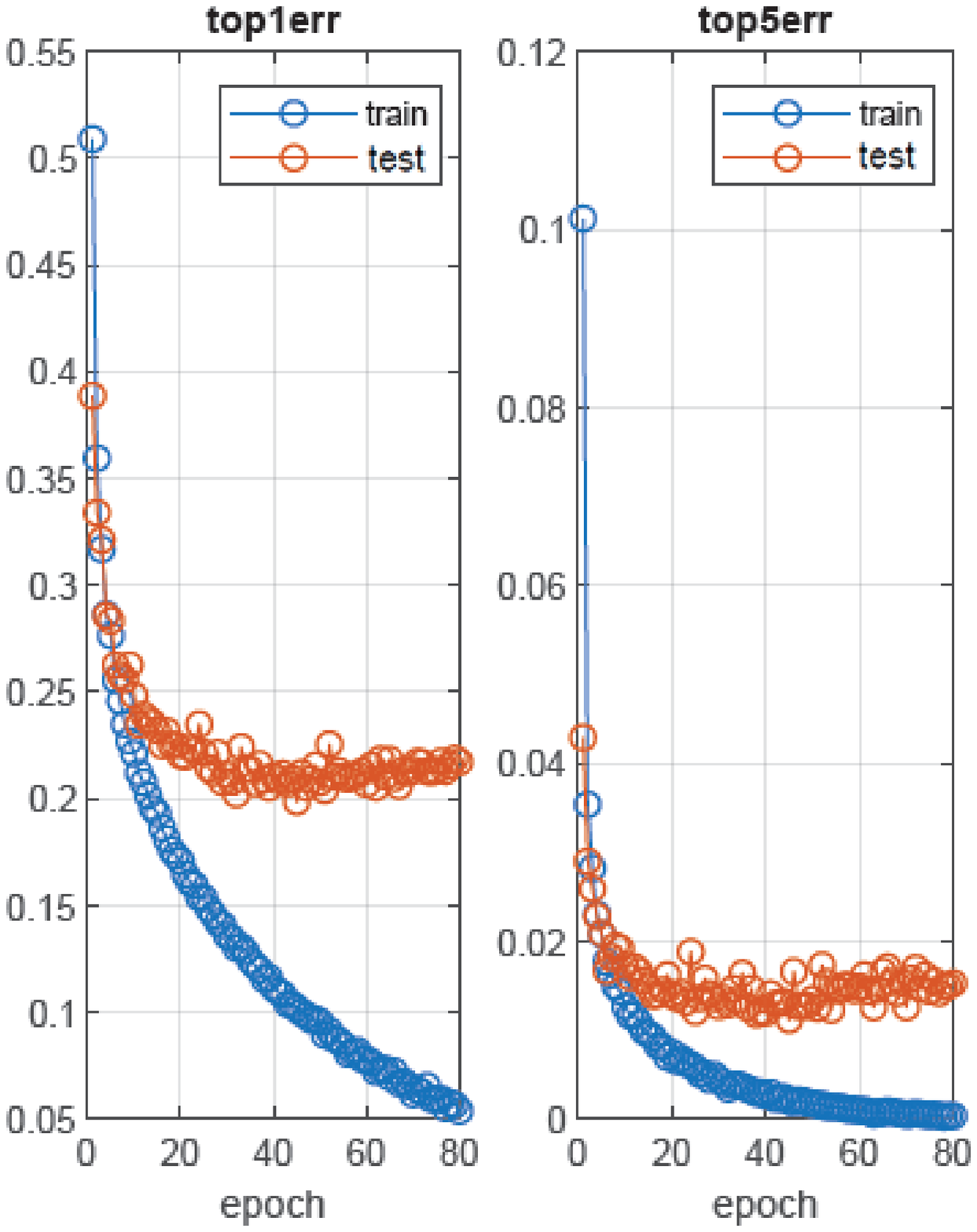}
	\includegraphics[width = 0.32\textwidth,height = 0.23 \textheight]{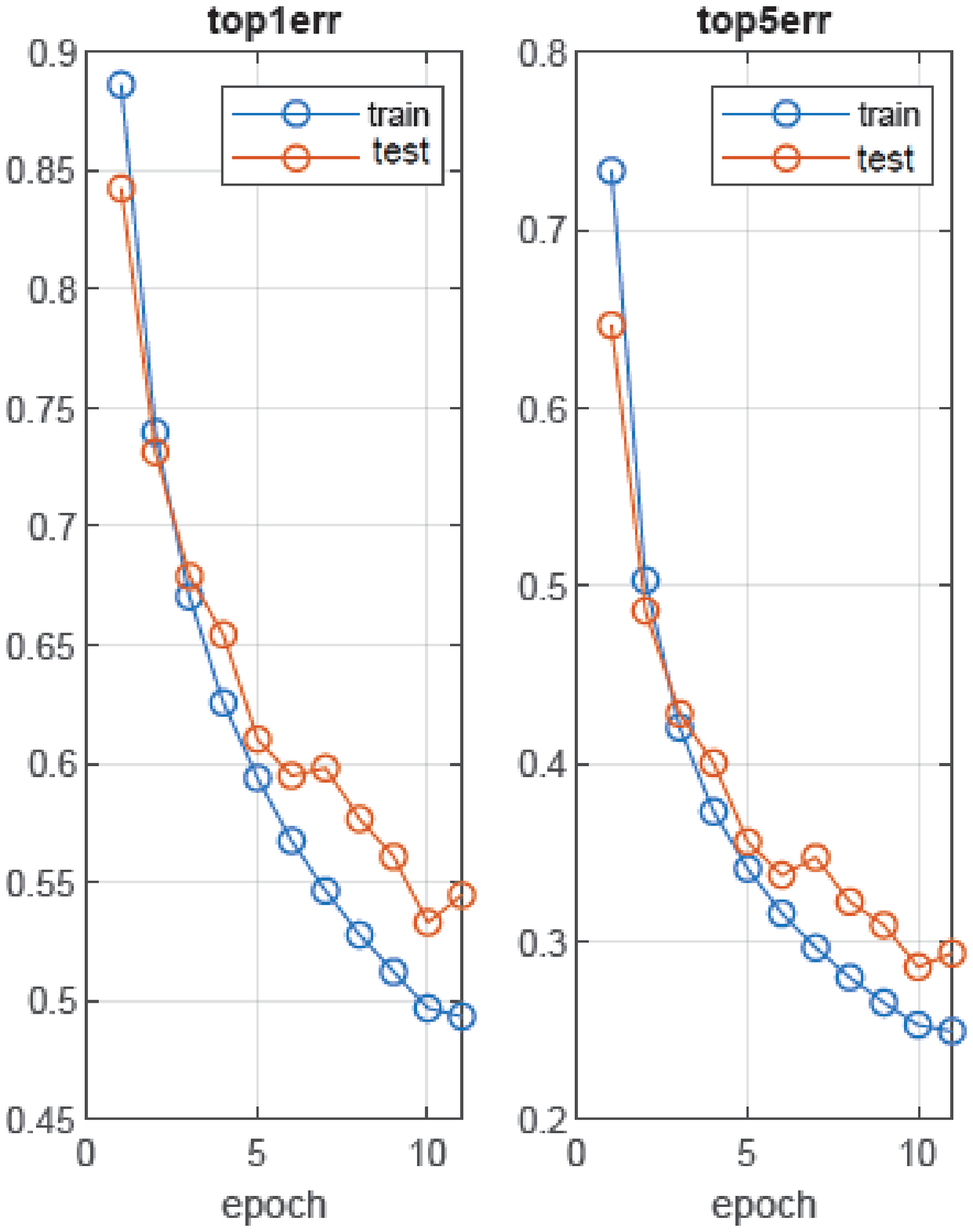}
	\includegraphics[width = 0.32\textwidth,height = 0.23 \textheight]{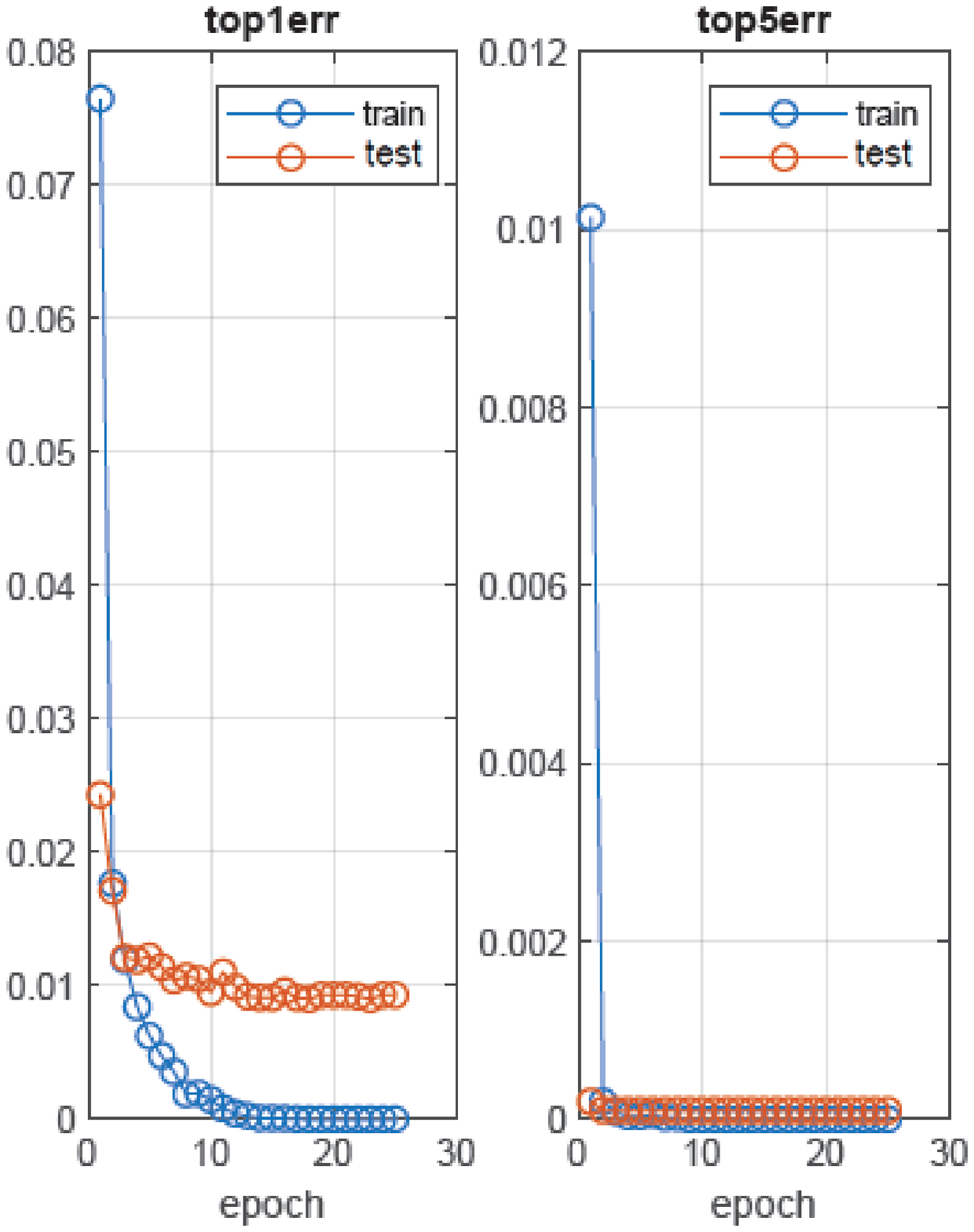}\\
	\includegraphics[width = 0.32\textwidth,height = 0.23 \textheight]{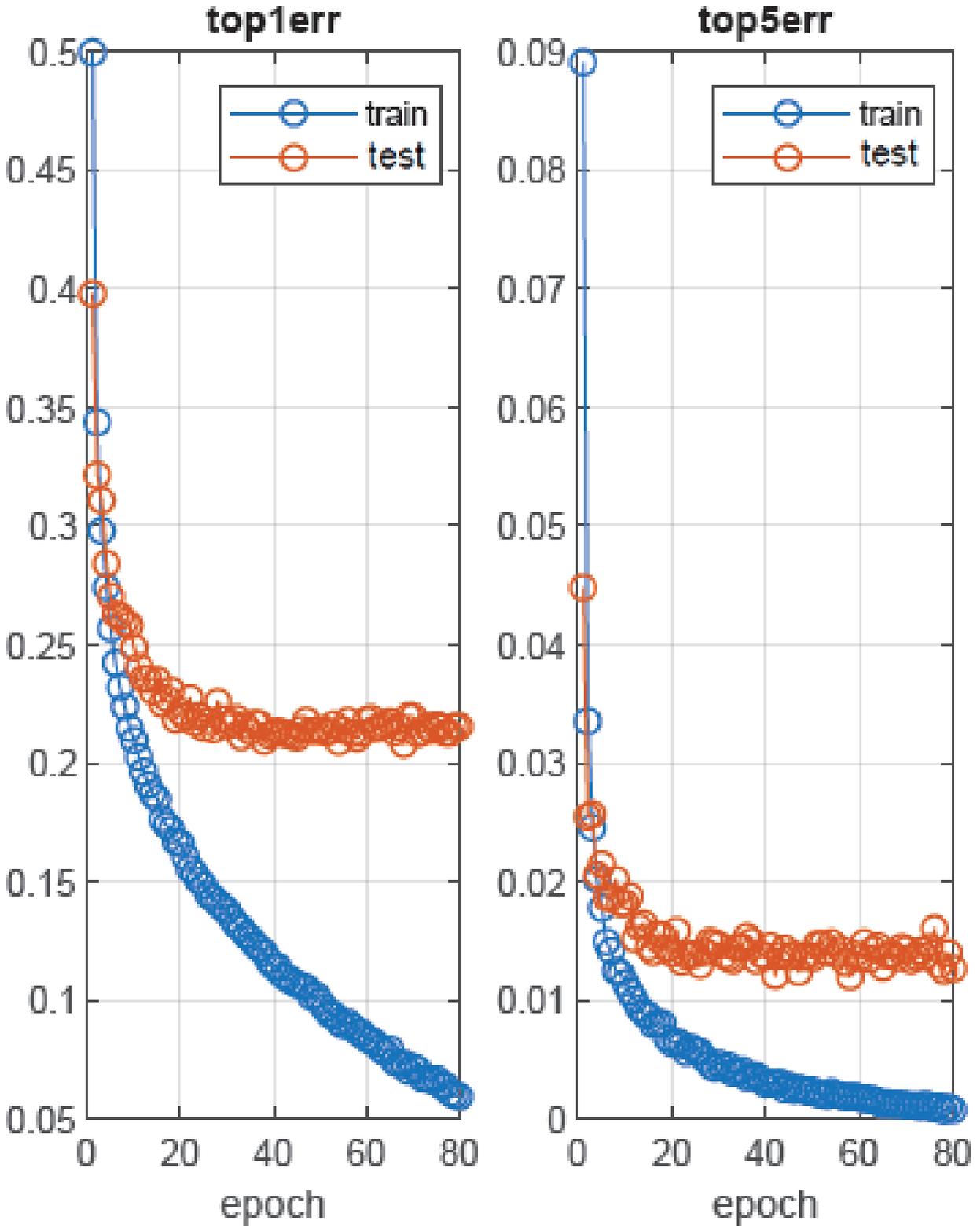} 
	\includegraphics[width = 0.32\textwidth,height = 0.23 \textheight]{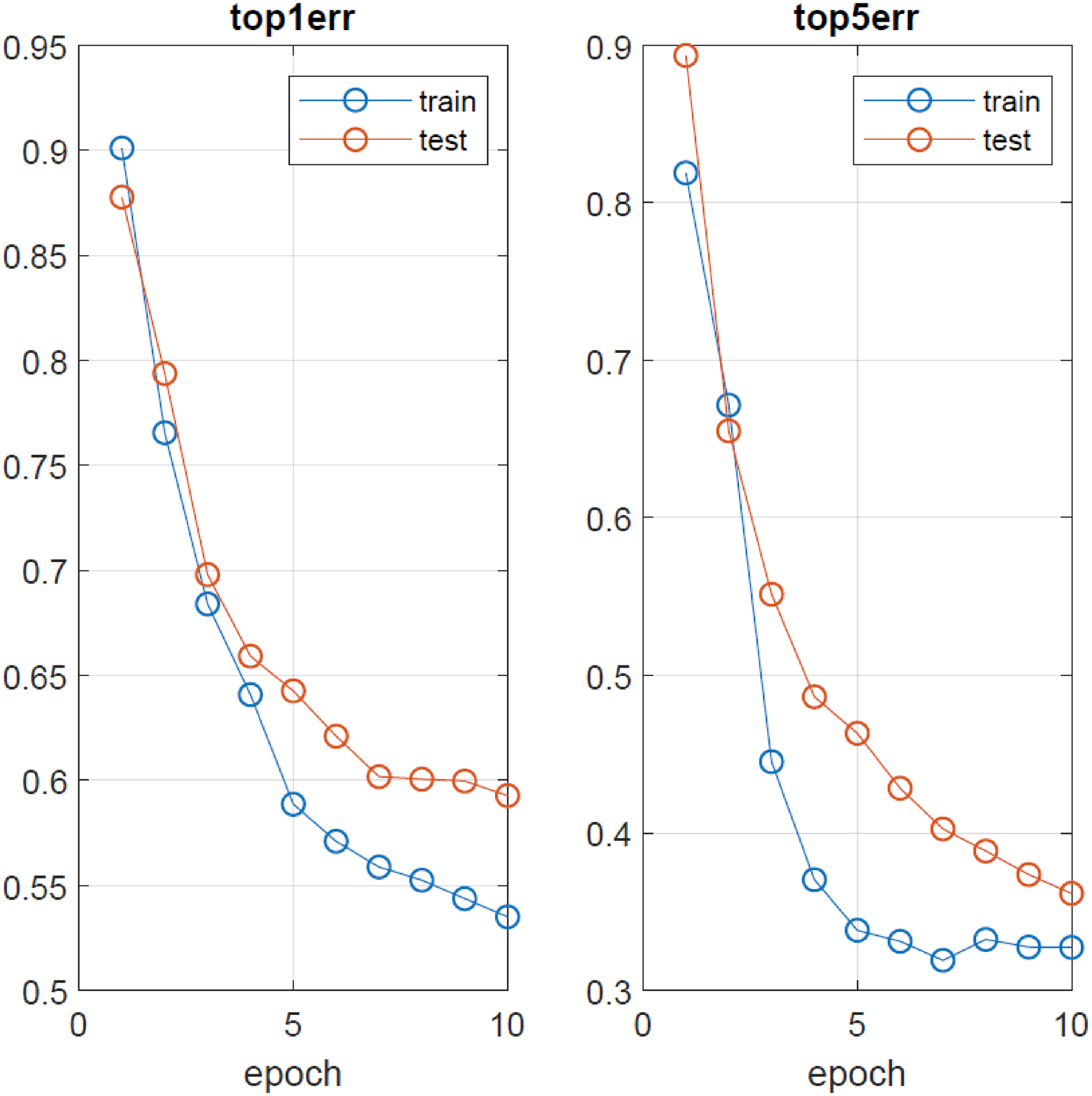}
	\includegraphics[width = 0.32\textwidth,height = 0.23 \textheight]{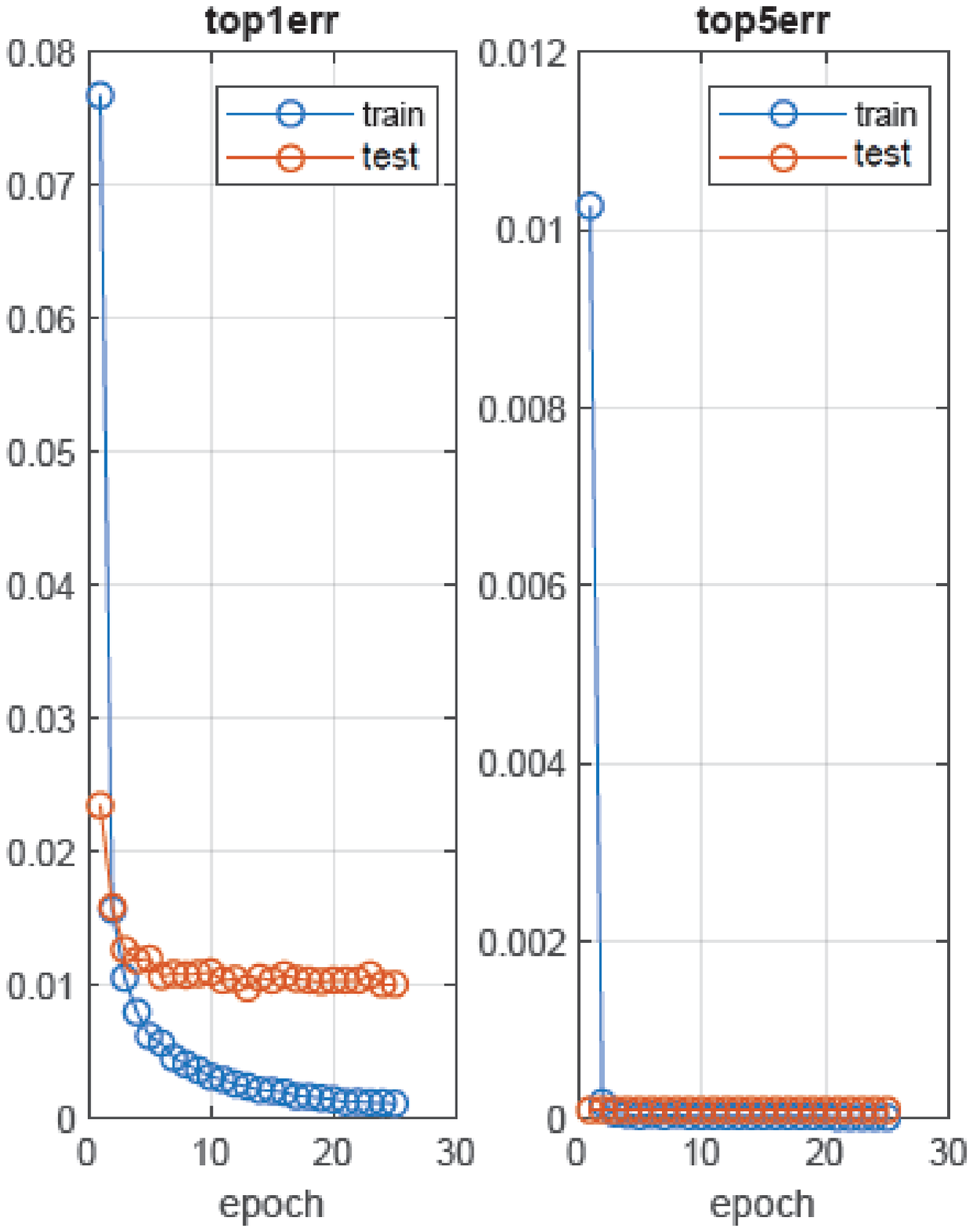}\\ 
	\includegraphics[width = 0.32\textwidth,height = 0.23 \textheight]{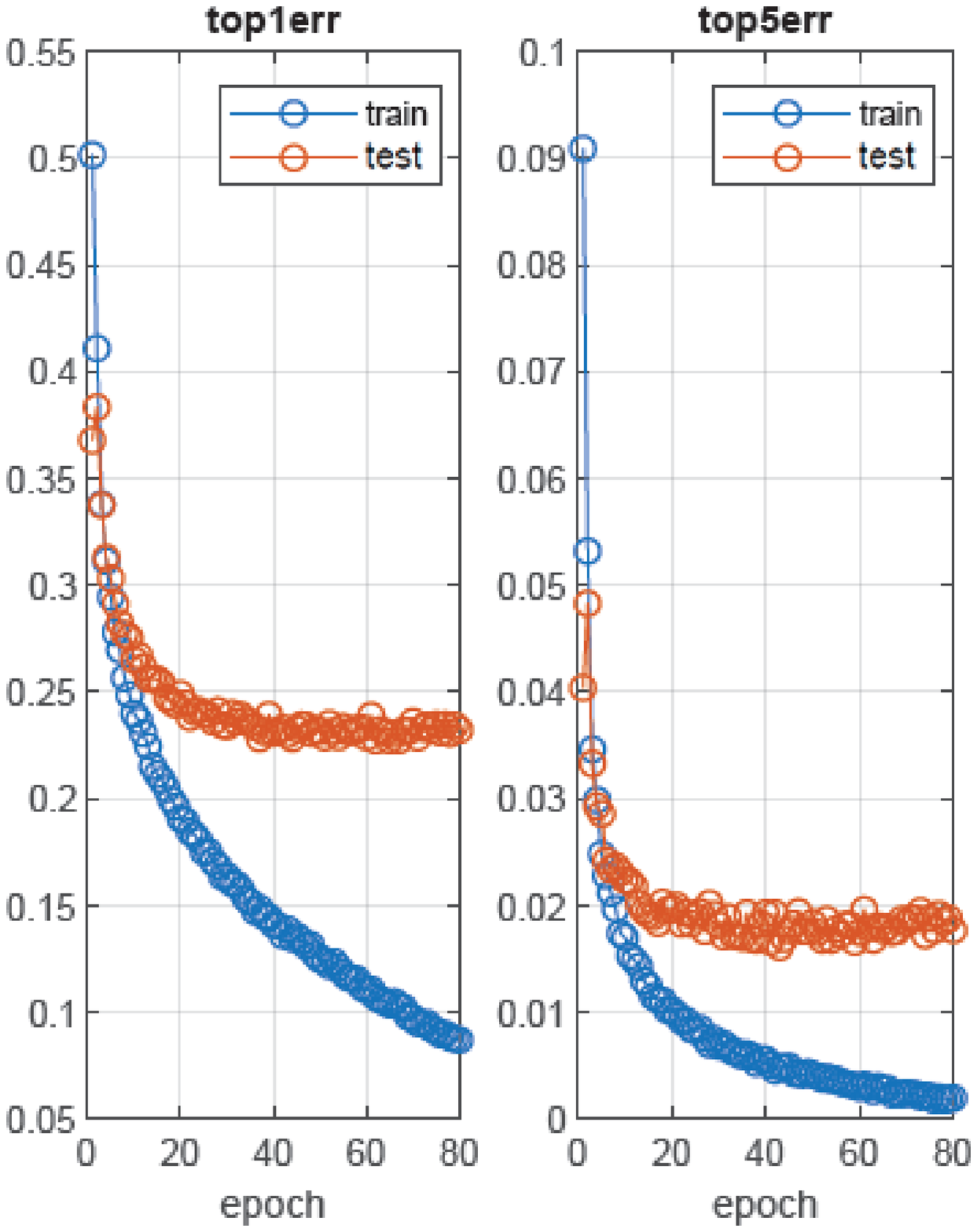}
	\includegraphics[width = 0.32\textwidth,height = 0.23 \textheight]{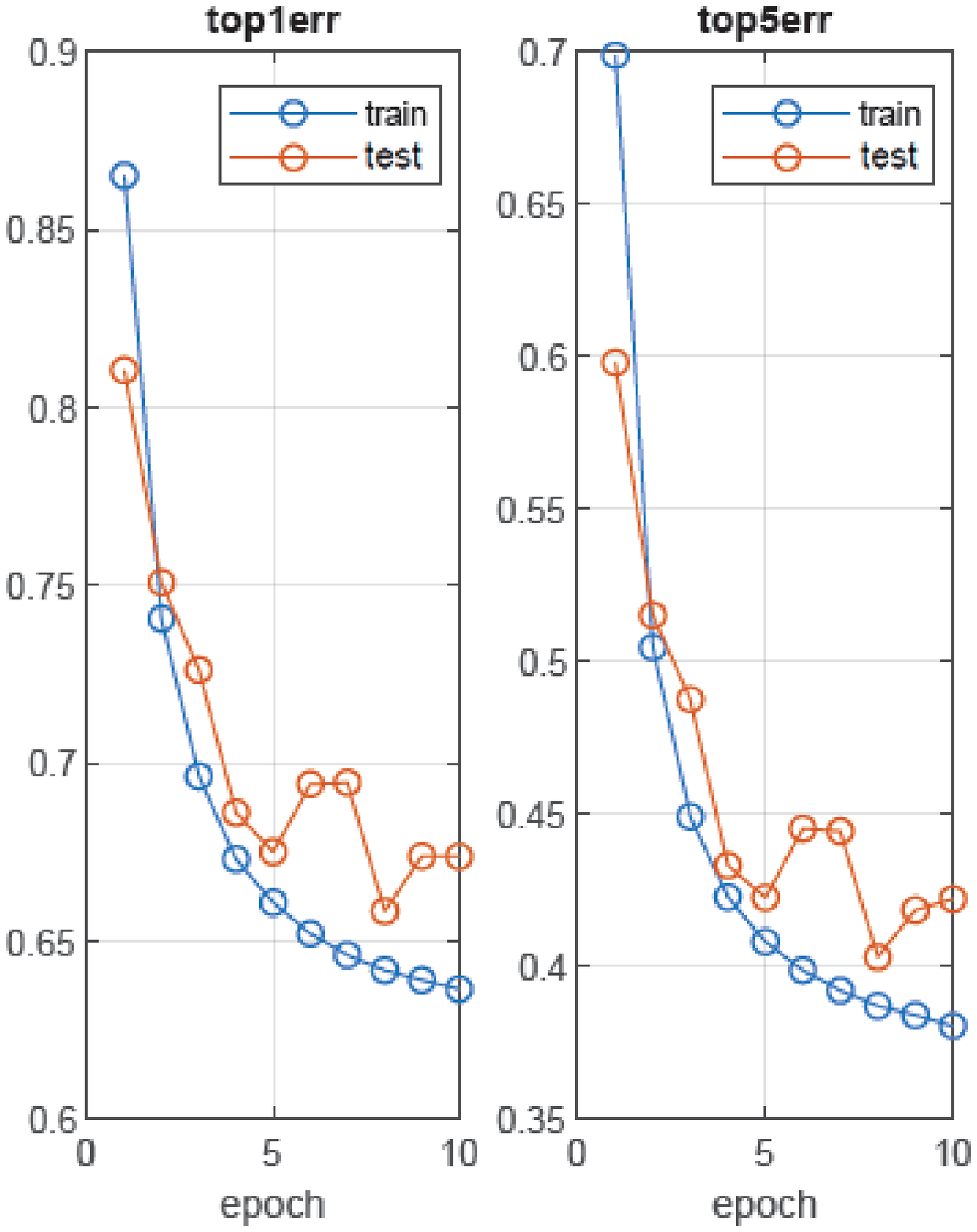}
	\includegraphics[width = 0.32\textwidth,height = 0.23 \textheight]{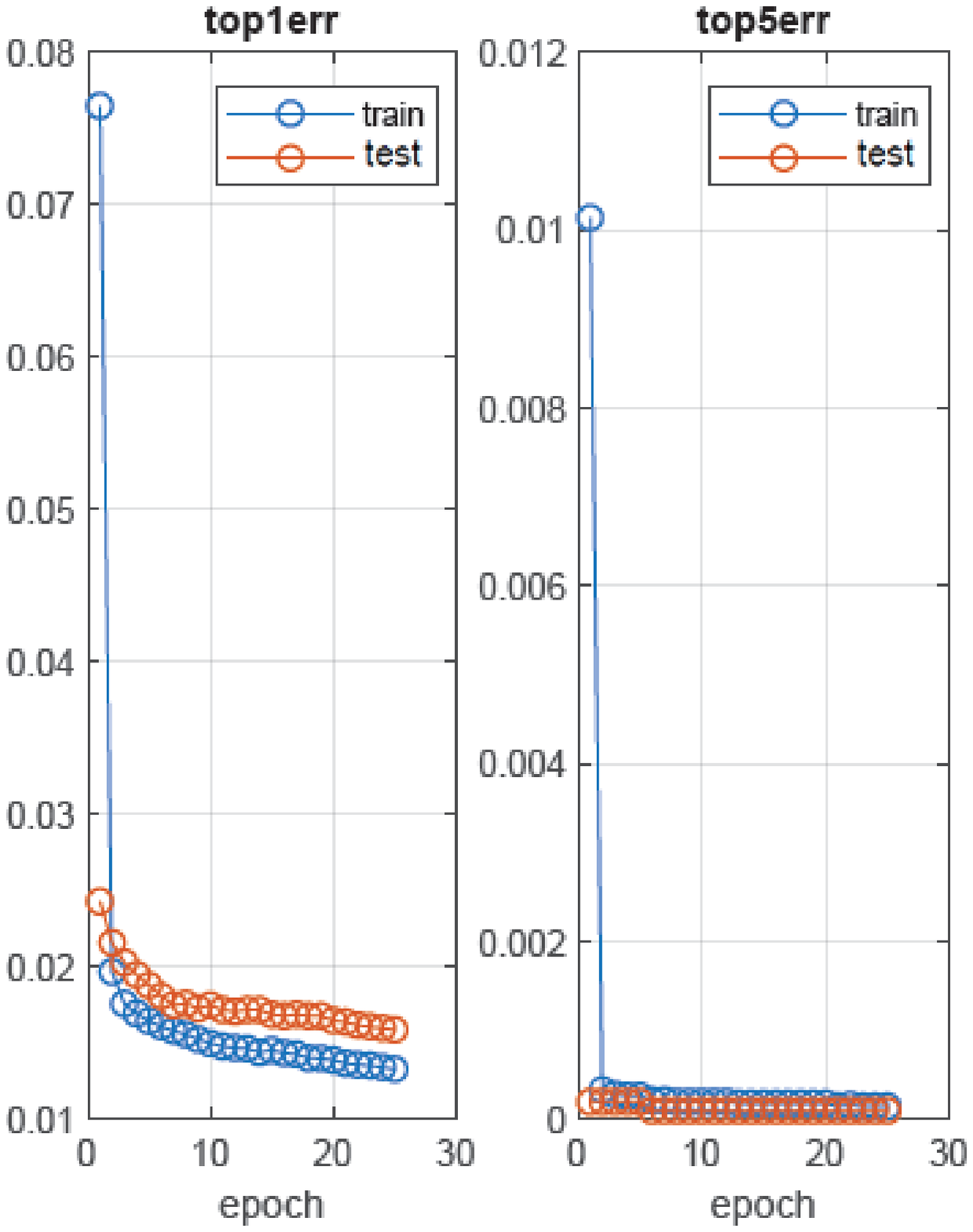}
	\caption{Error as a function of epoch, {\it i.e.} iteration, number for the three datasets under study obtained using, from top-to-bottom, our Barzilai-Borwein learning rate, Adam\cite{dozat:2016}, Adadelta\cite{zeiler:2018} and the baseline. From left-to-right, the columns correspond to CIFAR10, Imagenet and MNIST.}
	\label{fig:01}
\end{figure*}

In Figure \ref{fig:01} we show the performance for the training and testing per epoch for each of the datasets under study when our approach, Adam\cite{dozat:2016}, Adadelta\cite{zeiler:2018} and a baseline is used. In the figure, from left-to-right, we show the first (top1err) and top-five (top5err) classification error, in decimal fractional percentage, for the CIFAR10, Imagenet and MNIST datasets. From top-to-bottom, the rows show the results yielded by our method, Adam\cite{dozat:2016}, Adadelta\cite{zeiler:2018} and a baseline where the learning rate is set using the formula in Equation \ref{eq:07}. The parameter $b$ for the baseline has been found by cross validation and set to $0.02$, $0.05$ and $0.001$ for CIFAR10, Imagenet and MNIST, respectively.

Note that, in the plots in Figure \ref{fig:01} we show both, the training and testing performance. The training performance per epoch is denoted ``train'' (plotted in a solid blue line) whereas the testing is denoted ``test'' (plotted in a red, solid line). This should not be confused with the use of a validation data split, rather this is the evaluation of the network on the testing set at the corresponding epoch. In Table \ref{tab:01} we show the performance results corresponding to the last epoch of the ``test'' plot in percentage. The table shows the percentage classification error rates for the three datasets and the four learning rate methods under study, with the best performing method indicated in bold fonts.

\begin{table}[b]
\caption{Testing error rate for each of the learning rate methods under consideration for the CIFAR10, Imagenet and MNIST datasets. Best performance is in bold font.\vspace{-0.4cm}}
\begin{center}
\normalsize{
\begin{tabular}{|l|c|c|c|}
\hline
\textbf{Learning rate}&\multicolumn{3}{|c|}{\textbf{Dataset}} \\
\cline{2-4} 
\textbf{method} & CIFAR10 & Imagenet & MNIST  \\
\hline
Ours (Barzilai-Borwein)& \bf{20.83\%} &  58.91\% &  0.97\%  \\
Adam\cite{dozat:2016} &  21.85\%  &  \bf{56.12\%} &  \bf{0.93\%}  \\
Adadelta\cite{zeiler:2018} &  21.17\% & 59.37\%  &  0.98\%  \\
Baseline (see text) &  24.22\% &  65.85\% &  1.58\% \\
\hline
\end{tabular}}
\label{tab:01}
\end{center}
\end{table}

\section{Discussion}

Note that, from Figure \ref{fig:01}, we can conclude that our method based on the Barzilai and Borwein \cite{barzilai:88} learning rate provides a smoother and often faster training convergence. This can be appreciated better in the cases of the plots for CIFAR10 and Imagenet. In terms of performance, from Table \ref{tab:01} we can see that our method still delivers better performance in the case of CIFAR10, being the second for Imagenet and MNIST with a negligible loss of performance with respect to Adam\cite{dozat:2016}. Moreover, note that our method tends to over fit less. This is particularly evident in the plots correspondign to CIFAR10, were Adam\cite{dozat:2016} and Adadelta\cite{zeiler:2018} do overfit towards the end of the the training process.

As mentioned earlier, Algorithm \ref{alg:01} can employ both, Adagrad\cite{duchi:2011} or RMSprop in Line 3. Moreover, the initialisation choice made here, to call Adam\cite{dozat:2016} could be equally substituted with Adadelta\cite{zeiler:2018}. In our experience with the three datasets, Adam provides a marginal initialisation improvement over Adadelta\cite{zeiler:2018}. The same can be said of Adagrad\cite{duchi:2011} which delivers a marginal improvement on the testing results as compared to RMSprop.

It is also worth noting that Adagrad\cite{duchi:2011} does not employ momentum\cite{qian:99} for the update of the parameters. This can be done by making use of a weighted cummulative of the subgradients as shown in Equations \ref{eq:10} and \ref{eq:11}. However, we have opted not to do so since momentum terms introduce the parameter $\beta$ which modifies the step size. Moroever, adding momentum may be viewed as equivalent to employing gradient descent with a re-scaled step size\cite{yuan:2016}. However, momentum parameters have been shown to improve results significantly in some cases\cite{bengio:2013} and remain an attractive option which we aim to explore further in the future. 

\section{Conclusions}

In this paper, we have shown how the Barzilai-Borwein \cite{barzilai:88} adaptive step size can be used for training deep neural networks using Adagrad\cite{duchi:2011}. Departing from the relationship between the learning rate and the backpropagation scheme used in neural networks, we illustrate how the two-point secant equation used in quasi-Newton optimisation methods can be used to obtain the Barzilai-Borwein adaptive step size as applied to deep network training. The method is quite general in nature and can be applied to other training methods based upon gradient descent such as RMSprop. We have shown results using standard network architectures and datasets. We have also compared our approach to alternatives elsewhere in the literature. In our experiments, our method delivers better and smoother convergence with comparable performance to that delivered by the alternatives. 

\section*{Acknowledgment}

The authors would like to thank NVIDIA for providing the GPUs used to obtain the results shown in this paper through their Academic grant programme.


\bibliographystyle{IEEEtranS}
\bibliography{./references}
\end{document}